\pdfoutput=1

\documentclass[11pt]{article}

\usepackage[preprint]{acl}

\usepackage{times}
\usepackage{latexsym}
\usepackage[utf8]{inputenc}
\usepackage[T1]{fontenc}
\usepackage{microtype}
\usepackage{inconsolata}
\usepackage{amsmath}
\usepackage{amsfonts}
\usepackage{amssymb}
\usepackage{listings}
\usepackage{graphicx}
\usepackage{subcaption}

\usepackage{booktabs}
\usepackage{multirow}
\usepackage{nicematrix}
\usepackage{pifont}
\usepackage{xcolor}
\usepackage[normalem]{ulem}
\usepackage[most]{tcolorbox}
\usepackage{makecell}
\usepackage{worldflags}
\usepackage{ulem}
\usepackage{makecell}

\usepackage{courier}
\usepackage{cuted}

\newcommand{\SwissADIcon}{\includegraphics[width=1.5em]{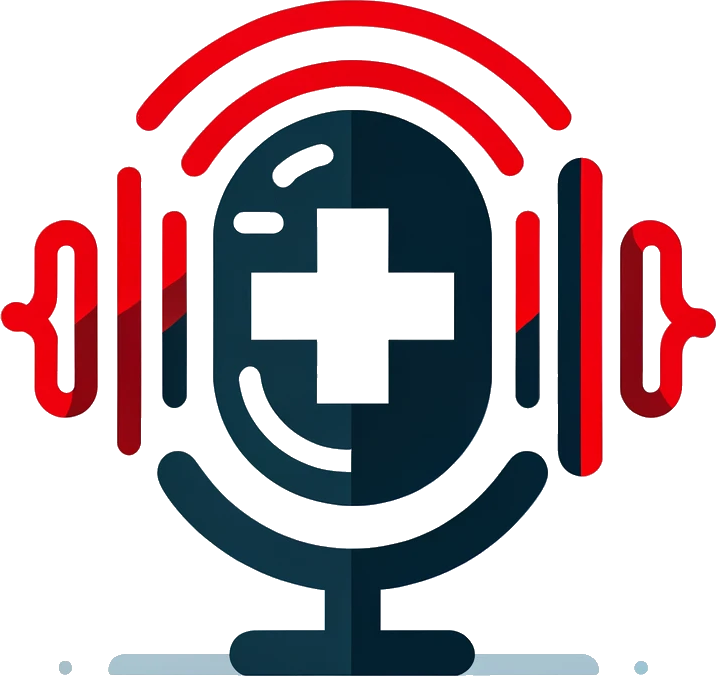}}

\newcommand{\uzhlogo}{\includegraphics[width=1.3em]{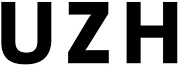}}
\newcommand{\zhawlogo}{\includegraphics[width=1em]{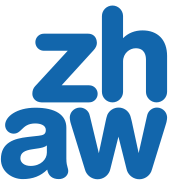}}

\UseRawInputEncoding

\title{\SwissADIcon \hspace{0.2em} SwissADT: An Audio Description Translation System for \\ Swiss Languages}

\author{$\textbf{Lukas Fischer}^{\uzhlogo}$, 
$\textbf{Yingqiang Gao}^{\uzhlogo}$,
$\textbf{Alexa Lintner}^{\zhawlogo}$, $\textbf{Sarah Ebling}^{\uzhlogo}$
\vspace{0.8em}
\\
$^{\uzhlogo}\text{Department of Computational Linguistics, University of Zurich, Switzerland}$ \\ 
\texttt{\{fischerl, yingqiang.gao, ebling\}@cl.uzh.ch} \\
$^{\zhawlogo}\text{School of Applied Linguistics, Zurich University of Applied Sciences, Switzerland}$ \\
\texttt{alexa.lintner@zhaw.ch} \\
 }

\begin{document}
\maketitle

\begin{abstract}
Audio description (AD) is a crucial accessibility service provided to blind persons and persons with visual impairment, designed to convey visual information in acoustic form. 
Despite recent advancements in multilingual machine translation research, the lack of well-crafted and time-synchronized AD data impedes the development of  audio description translation (ADT) systems that address the needs of multilingual countries such as Switzerland. 
Furthermore, since the majority of ADT systems rely solely on text, uncertainty exists as to whether incorporating visual information from the corresponding video clips can enhance the quality of ADT outputs.
In this work, we present SwissADT, the first ADT system implemented for three main Swiss languages and English. By collecting well-crafted AD data augmented with video clips in German, French, Italian, and English, and leveraging the power of Large Language Models (LLMs), we aim to enhance information accessibility for diverse language populations in Switzerland by automatically translating AD scripts to the desired Swiss language. Our extensive experimental ADT results, composed of both automatic and human evaluations of ADT quality, demonstrate the promising capability of SwissADT for the ADT task. We believe that combining human expertise with the  generation power of LLMs can further enhance the performance of ADT systems, ultimately benefiting a larger multilingual target population.\footnote{Our system is hosted on GitHub: \url{https://github.com/fischerl92/swissADT} and running at \url{https://pub.cl.uzh.ch/demo/swiss-adt}. 
}

\end{abstract}

\section{Introduction}

AD denotes the process of acoustically describing relevant visual information that renders streaming media content in television or movies and other art forms partly accessible to blind persons and persons with visual impairment  \citep{bardini2020audio, wang2021toward, ye2024mmad}. This service involves the creation of textual descriptions, so-called ``AD scripts'', of key visual elements of a scene, such as actions, environments, facial expressions, and other important details that are not conveyed through dialogue, sound effects, or music \citep{snyder2005audio, mazur2020audio}. They are typically inserted into natural pauses that do not interfere with the ongoing narration. AD scripts are  voiced by a  professional human speaker or synthesized by a computer and mixed with the original audio.

Despite recent advancements in multilingual machine translation \citep{liu2020multilingual, xue2021mt5} and Large Language Models (LLMs) research \citep{brown2020language, achiam2023gpt}, two major challenges remain unsolved in developing well-performing ADT systems. Firstly, many ADT systems are built on pre-trained machine translation models that need texts in both the source and target languages as inputs. Training these ADT systems requires large amounts of manually crafted data, leading to high operational costs \citep{ye2024mmad}. Secondly, existing ADT systems are predominantly text-only machine translation models, neglecting the visual modality which is  paramount for the ADT task and has proven to be useful as part of multimodal machine translation \citep{li2021vision}.  

In Switzerland, the primary target group of AD users comprises approximately 55,000 blind persons and 327,000 persons with visual impairment \citep{spring-2020}. Switzerland's multilingual population poses a challenge, as AD scripts have to be produced parallelly in several languages. This highlights the need of implementing multilingual ADT systems. 

\begin{figure*}
    \centering
    \includegraphics[width=\textwidth]{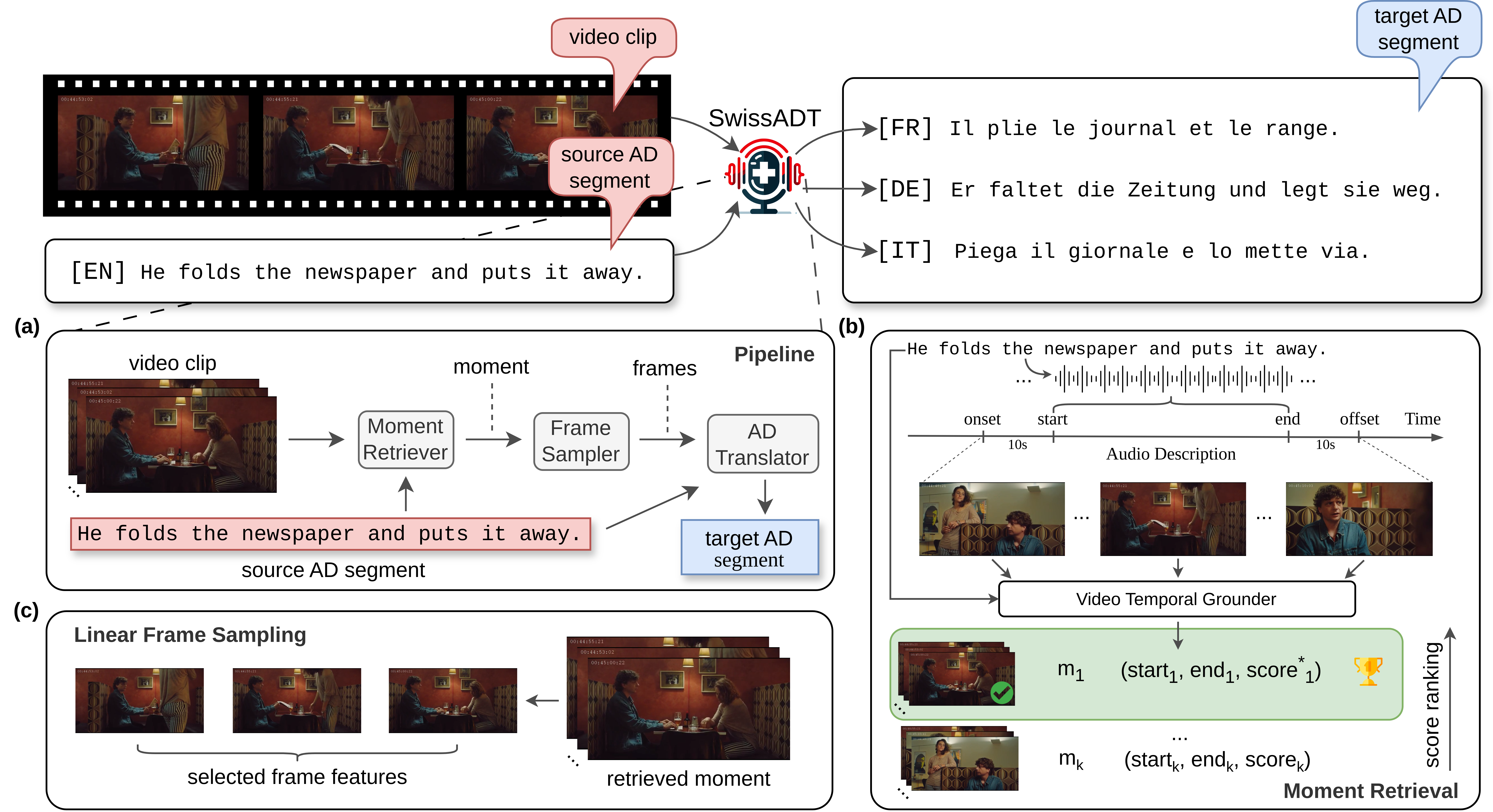}
    \caption{\textbf{(a) Overview of SwissADT}: An end-to-end pipeline that translates a given AD segment from English to the three main languages of Switzerland with the most salient video frames; \textbf{(b) Detail of the moment retriever}: it selects a moment, i.e., the most salient sequence of consecutive frames, to augment the translation inputs; (c) \textbf{Detail of the frame sampler}: it linearly interpolates the retrieved moment to obtain a cascade of frames used as inputs to the \textbf{AD translator}. In our implementation, we choose LLMs (GPT-4 models) as the AD translator due to their superior capabilities for performing multilingual machine translation tasks.}
    \label{fig:pipeline}
\end{figure*}

In this work, we  address the aforementioned challenges by developing an ADT system specifically for the three main languages of Switzerland, i.e., German, French, and Italian. To create training data for LLM-based ADT models with minimal human effort, we utilize DeepL\footnote{\url{https://www.deepl.com/de/translator}} with English as an auxiliary language to generate AD scripts in the three Swiss languages. 
To verify if LLMs are a potential solution to ADT task, we conduct automatic and human evaluations of LLM-generated AD scripts. 
Additionally, to further improve the translation quality, we incorporate video clips as part of the inputs to the LLM-based ADT models. 

Our contributions are: 1) We propose SwissADT, the first multilingual and multimodal ADT system for Swiss languages; 2) We conduct extensive evaluations of our ADT systems using both automatic and human quality assessments; and 3) We provide the source code for SwissADT, which is easily installable for reproducibility.   

\section{Related Work}

The automatic generation of ADs from video clips has been explored by both the natural language processing (NLP) and computer vision (CV) communities. This research is often conducted as part of tasks such as video captioning (generating descriptive text for a video) or video grounding (temporally aligning a text query with video segments). 

\begin{table*}
    \centering
    \begin{tabular}{lrrrrr}
    \toprule
    Language & \# Files & \# Characters & Video Hours & AD Hours & Ratio \\
    \midrule
    German  & 144 & 1,197,254 & 144:24:52 & 20:07:25 & 13.93\% \\
    French & 30 & 569, 535 & 28:53:24 & 8:44:00 & 30.23\% \\
    Italian & 23 & 486, 135 & 26:57:59 & 9:18:47 & 34.54\% \\
    Swiss German & 95 & 945, 865 & 71:31:32 & 15:27:21 & 21.61\% \\
    \midrule
    total  & 292 & 3,168,789 & 271:47:48 & 53:37:33 & 19.73\%  \\
    \bottomrule
    \end{tabular}
    \caption{Overview of our aggregated AD data.}
    \label{tab:data-overview}
\end{table*}

In recent years, several datasets and models for ADs have been published, where many of them are movie subtitles or video descriptions \citep{chen2011collecting, lison2016opensubtitles2016, xu2016msr,  lison2018opensubtitles2018}. 
\citet{oncescu2021queryd} proposed QuerYD, an open-source dataset created for the text-video retrieval and event localization tasks, where ADs and video segments are annotated by human volunteers. 
\citet{soldan2022mad} presented MAD, a large-scale benchmark dataset for video-language grounding, aggregated by aligning ADs with their temporal counterparts in videos. \citet{zhang2022movieun} introduced MovieUN, a large benchmark specifically designed for the movie understanding and narrating task in Chinese movies. \citet{han2023autoad} released AutoAD, a model that leverages both text-only LLMs and multimodal vision-language models (VLMs) to generate context-conditioned ADs from movies. In another work of theirs \citep{han2023autoadii}, the authors further developed an extended model to address three crucial perspectives of AD generation, i.e., \textit{actor identity (who)}, \textit{time interval (when)}, and \textit{AD content (what)}. 
Despite benefiting from existing large-scale corpora and state-of-the-art research in NLP and CV, these works are limited to monolingual applications. Consequently, they fail to meet the needs of Switzerland's multilingual population.

A second line of research explores the feasibility and suitability of applying machine translation models for ADT which was originally conceived as a human task.
In the study conducted by \citet{fernandez2016machine}, the \textit{creation}, \textit{translation}, and \textit{post-editing} of English-Catalan AD script pairs were extensively investigated to assess whether machine-translated AD scripts achieved satisfactory quality. The authors found that machine translation models can serve as a feasible solution. \citet{vercauteren2021evaluating} studied English-Dutch AD script pairs and found that errors were prevalent in the machine-translated AD scripts, indicating that post-editing by human experts was necessary.

In contrast to some of the studies above, we show that introducing visual inputs to ADT systems can lead to improved results, as verified by our AD professionals during the human evaluation. 

\section{SwissADT: An ADT System for Swiss Languages}

SwissADT is a multilingual and multimodal LLM-based ADT system that translates AD scripts between English and three main languages of Switzerland with both visual and textual inputs. It contains three basic components:

\paragraph{Moment Retriever} To identify the most relevant moment (i.e., a sequence of consecutive frames) in a video clip for a given AD segment, we initially select a video segment that spans from ten seconds before the AD's start runtime (onset) to ten seconds after its end runtime (offset)\footnote{Adding ten-second buffers ensures that the described moment is fully included in the video segment. Although ADs are usually synchronized with the described content, they may be shifted in dialogue-heavy scenes to fit no-speech segments. This buffer, recommended by our AD experts, sufficiently captures the described content even with such shifts.}. We then apply the video temporal grounder CG-DETR \citep{moon2023correlation}, which takes in both the AD script and the selected video segment and outputs the most relevant moment of variable length by providing the start and end times, along with a grounding score. 
The final moment is retrieved by selecting the top-ranked moment with the highest grounding score from the pool of candidate moments. 

\paragraph{Frame Sampler} We linearly sample multiple video frames from the retrieved moment.\footnote{Linear sampling reliably includes frames that are representative of the entire segment. We leave other sampling methods for future research.}
These frames are then utilized as visual inputs of the AD translator. We empirically report results on using four frames and every 50th frame.\footnote{In our system, the number of video frames can be manually set by the user.}

\paragraph{AD Translator} We deploy multilingual and multimodal LLMs as the backbone AD translator of SwissADT. We conduct experiments with the fundamental GPT-4 models \texttt{gpt-4o} and \texttt{gpt-4-turbo}. We decide to apply zero-shot learning as part of  a cost-effective solution.

Our modularized implementation of SwissADT streamlines the integration of state-of-the-art LLM research outcomes. This design allows for the seamless incorporation of cutting-edge moment retrievers and AD translators with minimal effort.

\section{Data Collection}

\subsection{AD Scripts and Video Clips}

We aggregate AD scripts from movies and TV shows that were aired on Swiss national TV stations, namely \textit{Schweizer Radio und Fernsehen} (SRF), \textit{Radio T\'el\'evision Suisse} (RTS), and \textit{Radiotelevisione Svizzera} (RSI). Table~\ref{tab:data-overview} gives an overview of the aggregated AD scripts.  

It is noteworthy that AD scripts in French and Italian occupy significantly more runtime in videos compared to those in German. 
This discrepancy arises from the data source: German ADs are predominantly derived from episodes of the TV game show \textit{1 gegen 100}, which features relatively static scenes (same studio setting and moderator throughout, with only the game candidates varying), thereby reducing the necessity for extensive ADs.
Conversely, French and Italian ADs are primarily sourced from movies and documentaries, which typically require more descriptive narration.

To facilitate the data storage, we use the SRT format (commonly used for subtitles) for ADs and mp4 format for videos. Figure~\ref{fig:AD-scripts} (Appendix~\ref{section:AD-script}) demonstrates an AD passage from our  dataset.  

\subsection{Synthetic ADs with DeepL}

Due to a lack of parallel data, we use DeepL to generate synthetic AD scripts for each language pair of our system. 

We translate all German, French, and Italian AD scripts into the other two Swiss languages, respectively, as well as into English. We include English as a mediating language in our ADT models to allow potential synergies with an AD script generation system developed by a research partner in our project. In addition, the moment retriever CG-DETR was trained on an English dataset, therefore, English is required as an intermediary language in our pipeline. For each source language, we randomly split the synthetic ADs into train, dev, and test sets (see Table~\ref{tab:dataset-split} for more detail). We limit the number of ADs in both the dev and test sets to 200 samples each to preserve training data for further experiments, given the 7,500-sample size for French and Italian. AD data is scarce, so we carefully balanced its usage between training and testing. Additionally, we maintained consistent sizes across all languages to ensure uniform evaluation.

\begin{table}[!htb]
    \centering
    \begin{tabular}{ccrr}
    \toprule
     Source & Split & \# ADs & \# Characters  \\
     \midrule
     \multirow{3}{*}{German} & train & 21,272 & 1,175,412 \\
     & dev & 200 & 10,648 \\
     & test & 200 & 11,194 \\
     \midrule
     \multirow{3}{*}{French} & train & 7,099 & 538,063 \\
     & dev & 200 & 15,533 \\
     & test & 200 & 15,939 \\
     \midrule
     \multirow{3}{*}{Italian} & train & 7,108 & 460,235 \\
     & dev & 200 & 13,332 \\
     & test & 200 & 12,568 \\
     \bottomrule
    \end{tabular}
    \caption{Dataset split for AD scripts of each source language. We use test sets for automatic ADT evaluation.}
    \label{tab:dataset-split}
\end{table}

We exclude Swiss German AD scripts due to the inadequate translation quality when using DeepL.

\section{Evaluation Method}

\subsection{DeepL Translation Quality Estimation}

We assess the quality of silver-standard AD scripts translated by DeepL using GEMBA-MQM \citep{kocmi2023gemba}, an LLM-based metric that employs three-shot prompting with GPT-4 to identify and annotate error spans. This evaluation is conducted on test sets comprising 200 ADs for each source-target language pair, with weights assigned to \textit{No Error}, \textit{Minor Error}, \textit{Major Error}, and \textit{Critical Error} being 0, 1, 5, and 10, respectively. Table~\ref{tab:deepl-assessment} presents the overall error weights of the DeepL-translated AD scripts. 

\begin{table}[!htb]
\centering
\begin{tabular}{lrrrr}
\toprule
& EN & DE & FR & IT \\
\midrule
 DE & \textbf{1.775} & - & 2.465 & 2.925  \\
FR & \textbf{1.585} & 3.295 & - & 3.075 \\
IT & \textbf{2.375} & 3.525 & 3.815 & - \\
\bottomrule
\end{tabular}
\caption{Quality estimation of the synthetic ADs generated by DeepL. Source languages are placed row-wise and target languages column-wise. All weights are below 4, indicating that translation errors do not exceed the major level requiring extensive modifications.}
\label{tab:deepl-assessment}
\end{table}

These results indicate that the errors in DeepL-translated AD scripts range from minor to major, which are considered acceptable from the machine learning perspective.

\renewcommand\theadalign{bc}
\renewcommand\theadfont{\bfseries}
\renewcommand\theadgape{\Gape[4pt]}
\renewcommand\cellgape{\Gape[4pt]}

\begin{table*}
    \centering
    \resizebox{\textwidth}{!}{
    \begin{tabular}{lcccccccccccc}
    \toprule
    \multirow{2}{*}{\makecell{AD \\ Translator}} & \multirow{2}{*}{\makecell{Input \\ Modality}} & \multicolumn{3}{c}{EN $\rightarrow$ DE} & & \multicolumn{3}{c}{EN $\rightarrow$ FR} & & \multicolumn{3}{c}{EN $\rightarrow$ IT} \\
    \cmidrule{3-5}
    \cmidrule{7-9}
    \cmidrule{11-13}
    & & \textsc{bleu} & \textsc{meteor} & \textsc{chrF} & & \textsc{bleu} & \textsc{meteor} & \textsc{chrF} & & \textsc{bleu} & \textsc{meteor} & \textsc{chrF} \\
    \midrule
    \texttt{gpt-4o} & text-only & 56.95 & 80.44 & 77.20 & & 65.75 & 83.58 & 80.74 & & \textbf{63.30} & 79.03 & \textbf{78.66} \\
    \texttt{gpt-4-turbo} & text-only & 54.27 & 78.08 & 76.10 & & 64.42 & 82.95 & 80.36 & & 58.64 & 77.94 & 76.29 \\
    \midrule
    \texttt{gpt-4o} & text + 4 frames & \textbf{58.20} & \textbf{81.23} & \textbf{78.20} & & \textbf{66.10} & 83.37 & \textbf{81.12} & & 63.15 & 79.24 & 78.31 \\
    \texttt{gpt-4o} & text + $n$ frames & 57.88 & 80.15 & 77.20 & & 65.59 & 83.40 & 80.75 & & 62.67 & \textbf{79.75} & 78.51 \\
    \texttt{gpt-4-turbo} & text + 4 frames & 54.61 & 77.47 & 75.80 & & 64.40 & \textbf{83.70} & 80.60 & & 57.99 & 77.40 & 76.20 \\
    \texttt{gpt-4-turbo} & text + $n$ frames & 54.06 & 78.21 & 76.00 & & 65.85 & 83.41 & 80.90 & & 58.58 & 77.99 & 76.21 \\
    \bottomrule
    \end{tabular}
    }
    \caption{Results of ADTs, where we highlight the best scores per system in bold. In the table, $n$ represents the number of frames sampled at intervals of every 50 frames. Consequently, $n$ varies depending on the duration of the retrieved moment (the average values of $n$ are: EN$\rightarrow$DE: 2.40, EN$\rightarrow$FR: 3.48, EN$\rightarrow$IT: 2.87).}
    \label{tab:ADT-results}
\end{table*}

\subsection{Automatic ADT Evaluation}

We use \textsc{bleu} \citep{papineni2002bleu}, \textsc{meteor} \citep{banerjee2005meteor}, and \textsc{chrF} \citep{popovic2015chrf} as automatic evaluation metrics for AD scripts translated by SwissADT, where the scores are calculated by comparing the generated AD scripts to the ground truths.

\subsection{Human Evaluation with AD Professionals}

We conduct human evaluations with our AD experts\footnote{We plan to gather feedback from visually impaired users in the future, once SwissADT reaches a sufficient quality level.} to assess the quality of AD scripts translated by SwissADT. 
Our objective is to verify the hypotheses that automatic evaluation scores reflect the human judgments well, and that multimodal inputs improve translation quality.

We utilize Microsoft Forms\footnote{\url{https://forms.office.com}} to conduct our study. Following the Scalar Quality Metric (SQM, \citet{freitag2021experts}) evaluations, we assess each AD pair (both source and target languages) along three dimensions: \textit{fluency}, \textit{adequacy}, and AD \textit{usefulness}. AD experts rate these dimensions on a seven-point scale (0 to 6). 
The assessment is conducted online, and we compensate the AD experts at a  rate of 85 CHF per working hour. We compare the translations of our best AD translator, \texttt{gpt-4o}, for two input modalities: text-only, and text with four frames as inputs for this assessment.

Due to the difficulties in hiring AD experts with sufficient knowledge of English for French and Italian, we concentrate on evaluating German AD scripts.
We thus recruit three AD experts A, B, and C for human evaluation. 
Feedbacks from the pre-study questionnaire show that all three AD experts have a degree in translation as well as professional experience ranging betweeen three to five to over ten years. Furthermore, AD experts B and C are also professionally trained post-editors.


For the human evaluation, we randomly sample 30 consecutive blocks of 10 AD segments from our German dataset. We choose consecutive AD segments so AD experts have more context to judge the translations.
To minimize bias, each AD expert evaluates the same 30 blocks, in randomized order. 

We use \texttt{gpt-4o} to translate the English silver AD segments back to German. We randomly select one of two strategies for each segment: text-only and text + four frames. The AD experts are presented with the English source segment and the German translation of \texttt{gpt-4o}, without knowing which input modality was used for the translations.

 We report weighted Cohen's kappa \citep{cohen1968weighted} for inter-evaluator agreement.


\section{Results and Discussions}

\subsection{AD Translations}

Table~\ref{tab:ADT-results} presents the automatic evaluations of various AD translators. We observe that
\begin{itemize}
    \item \texttt{gpt-4o} outperforms \texttt{gpt-4-turbo};
    \item GPT-4-based results demonstrate promising performance in the ADT task, as indicated by high evaluation scores. This finding supports the effectiveness of applying machine translation models to address the ADT task, which is aligned with previous literature;
    \item Augmenting source ADs with corresponding video frames generally enhances translation quality, with the inclusion of more input frames leading to improved results. This suggests that it is beneficial to incorporate the visual modality into the ADT pipeline to utilize the power of fundamental LLMs.
\end{itemize}

The slightly better performance of \texttt{gpt-4o} with text-only on EN$\rightarrow$IT may be due to language-specific factors, the small dataset size or varying multilingual zero-shot capabilities, as the differences are not statistically significant. This result does not undermine the hypothesis that multimodal input improves translation quality overall, as other language pairs show the expected benefits. For examples where visual input is beneficial, refer to Appendix~\ref{section:examples}.





\begin{table*}[!htb]
\begin{subtable}[t]{0.5\textwidth}
    \centering
    \begin{tabular}{lrrr}
    \toprule
    text-only & A\&B & B\&C & A\&C \\
     \midrule
     fluency & 0.30 & 0.22 & 0.21 \\
     adequacy & 0.38  & 0.25 &  0.33 \\
     usefulness & 0.21 & 0.18 & 0.35 \\
     \midrule
     text-only & A & B & C \\
     \midrule
     avg. fluency & 5.28 & 4.95 & 5.50 \\
     avg. adequacy & 5.53 & 5.74 & 5.77\\
     avg. usefulness & 5.18 & 5.38 & 5.76\\
    \bottomrule
    \end{tabular}
    \caption{AD translator with only texts as inputs.}
\end{subtable}
 \hfill
 \begin{subtable}[t]{0.5\textwidth}
     \centering
     \begin{tabular}{lrrr}
    \toprule
    text + 4 frames & A\&B & B\&C & A\&C \\
     \midrule
     fluency & 0.29 & 0.25 & 0.20 \\
     adequacy   & 0.35 &  0.40 & 0.39  \\
     usefulness & 0.14 & 0.38 & 0.18 \\
     \midrule
     text + 4 frames & A & B & C \\
     \midrule
     avg. fluency & 5.37 & 5.16 & 5.61\\
     avg. adequacy & 5.62 & 5.77 &  5.70\\
     avg. usefulness & 5.12 & 5.27 &  5.78\\
    \bottomrule
    \end{tabular}
    \caption{AD translator with 4 video frames as inputs.}
 \end{subtable}
    \caption{Pairwise inter-evaluator agreement scores on AD fluency, adequacy, and AD usefulness, measured with Cohen's weighted kappa \citep{cohen1968weighted}. We also report the average evaluation scores of individual AD experts.}
    \label{tab:kappa}
\end{table*}

Given that training human AD experts requires completing a curriculum that encompasses numerous essential competences and skills  \citep{matamala2007designing, jankowska2017blended, colmenero2019training}, there is a persistent shortage of AD experts available to AD producers.
Consequently, implementing automatic ADT systems based on multilingual and multimodal LLMs followed by human post-editing could leverage AD production.

\subsection{Human Evaluation}



Table~\ref{tab:kappa} presents the inter-evaluator agreement results conducted with our AD experts as well as the average evaluation scores given by each AD expert, respectively.
First, we see that our AD experts demonstrate a fair level of agreement overall, highlighting the inherent difficulty in evaluating AD translations even among professionally trained individuals. Given this subjective variability among human evaluators, we contend that automatic evaluation metrics remain essential, as they offer an additional objective assessment independent of the evaluators' training. 
We also observe that AD scripts translated with four frames as input are rated higher in fluency (i.e. 5.38), and adequacy (i.e. 5.70) as compared to the text-only input translations (fluency: 5.24, adequacy: 5.68). These results verify our hypothesis that multimodal input improves translation quality. The dimension AD usefulness, however, is rated slightly higher for the AD scripts translated with the text-only input (i.e. 5.44) as compared to the four-frames translations (i.e. 5.39). 
Given the subjective opinions of AD experts on the AD usefulness, we argue that involving end users may yield more objective evaluations. This approach will be explored in our future research.


\section{Conclusions and Future Work}

In this work, we present SwissADT, a multilingual and multimodal ADT system designed to support three Swiss languages and English. Our findings demonstrate that leveraging LLMs to address the ADT task represents a significant initial step towards achieving information accessibility, as validated by our experienced AD experts. We anticipate that our work will  benefit  blind persons and persons with visual impairments in Switzerland, enhancing their  access to streaming media.

In future work, we aim to explore other LLMs for the ADT task by fine-tuning these models using the training data. 
Moreover, we plan to integrate the ADT pipeline with human-in-the-loop principles, training the system with reinforcement learning approaches to better align outputs of ADT systems with human preferences. 
We believe that integrating human expertise into the LLM pipeline for the ADT task will more effectively meet end users' expectations and satisfaction. As with any accessibility technology, it is paramount that it serves the needs of the end users.

\section{Limitations}

The limitations of our work are the following: 1) Due to the lack of high-quality data, we do not include Romansh as a target AD language, despite it being an official language of Switzerland that has nearly 35,000 native speakers\footnote{Source: \href{https://www.bfs.admin.ch/bfs/de/home/statistiken/bevoelkerung/sprachen-religionen/sprachen.assetdetail.31025129.html}{Swiss Federal Statistical Office}}; 2) Given the difficulty in sourcing AD experts for French and Italian, we are unable to conduct human evaluations for these two languages. However, we expect the results to be comparable to German ADs, as indicated by the comparable translation results of our best AD translator \texttt{gpt-4o}; 3) The multimodal nature of ADs has not been taken into account in the human evaluation, which would require the AD experts to have access to the visual inputs;
4) We do not utilize the Swiss German part of our dataset, as the absence of standardized spelling rules in Swiss German still poses a challenge for machine translation systems. This is primarily due to the fact that each word in Swiss German can have multiple spelling variations, resulting in an expanded vocabulary size.
However, we recently learned that Swiss German to German translation models developed by TextShuttle\footnote{\url{https://www.textshuttle.com}} produces promising results, and we will integrate their models into the SwissADT pipeline.



\section{Ethics Statement}

To ensure privacy protection and data anonymization, we  formally obtained informed consent for data collection of human ratings as per the guidelines of the Zurich University of Applied Sciences. 

\section*{Acknowledgments}
This work was funded by the Swiss Innovation Agency (Innosuisse) Flagship Inclusive Information and Communication Technologies (IICT) under grant agreement PFFS-21-47. We thank our industry partners SWISS TXT and SRG, particularly, Daniel McMinn and Veronica Leoni, for providing us with the use case and making data available. 


\bibliography{custom}

\appendix 






\newpage

\section{Audio Description Scripts}
\label{section:AD-script}

We make use of a common format for subtitles, namely SRT, where we treat ADs as subtitles.  See Figure~\ref{fig:AD-scripts} for detailed data schema.

\begin{figure}[!htb]
\centering
\begin{tcolorbox}[
  width=\columnwidth,
  colback=white, 
  colframe=white, 
  fontupper=\ttfamily\small, 
  fontlower=\ttfamily\small 
]
  7 \\
  00:01:13,240 -> 00:01:16,720 \\
  \$ Eine wuchtige Rolls Roice Luxus-Limousine. * Ein H\"andler kommt: \\

  8 \\
  00:01:42,240 -> 00:01:45,360 \\
  Chris nickt l\"achelnd. \\
  \$\$ Der H\"andler \"offnet die Autot\"uren. \\

  9 \\
  00:01:46,200 -> 00:01:51,360 \\
  UT: Toll. Es gibt nicht viele Autos f\"ur so grosse Menschen wie mich. So viel Beinfreiheit.
\end{tcolorbox}
\caption{An example of a German AD script with spoken subtitles and special characters used in our data schema. The presence of a dollar sign (\$) signifies a constrained timeframe of faster pace of speech. An asterisk sign (*) indicates a scene change within the script. Spoken subtitles are marked by UT as an abbreviation for ``Untertitel'' in German.}
\label{fig:AD-scripts}
\end{figure}

\section{Pricing}

To estimate the cost of translating large datasets of ADs, we provide the calculations in Table~\ref{tab:pricing} based on our dataset. Notice that OpenAI's pricing policy is subject to change, and that other factors, such as resolution and size of the input frames, as well as frequency and length of AD segments have great influence on the total price. 

\begin{table}[!htb]
    \centering
\resizebox{\columnwidth}{!}{
\begin{tabular}{llrr}
\toprule
\multirow{2}{*}{\textbf{Model}} & \multirow{2}{*}{\textbf{Pricing}} & \multicolumn{2}{c}{\textbf{Cost for 190 ADs}} \\
\cmidrule{3-4}
 &  & \textbf{text-only} & \textbf{text + 4 frames} \\ \midrule
\multirow{3}{*}{\texttt{gpt-4o}} & 5.00 \$ / 1M input tokens & \$0.06 & \$4.28 \\
 & 15.00 \$ / 1M output tokens & \$0.06 & \$0.06 \\ \cmidrule{2-4} 
 & \textbf{total} & \textbf{\$0.11} & \textbf{\$4.33} \\ \midrule
\multirow{3}{*}{\texttt{gpt-4-turbo}} & 10.00 \$ / 1M input tokens & \$0.11 & \$8.55 \\
 & 30.00 \$ / 1M output tokens & \$0.11 & \$0.11 \\ \cmidrule{2-4} 
 & \textbf{total} & \textbf{\$0.23} & \textbf{\$8.66} \\ \bottomrule
\end{tabular} }
    \caption{Expected translation costs for an average AD script (assuming a video duration of 56 minutes, 190 AD segments). We resize the input frames to 960x540 pixels, which results in roughly 4,500 total input tokens (including text prompt) for a single ADT with 4 frames. The average lenght of text-only prompts is 60 tokens, and the average output length is 20 tokens.
    Pricings of \texttt{gpt-4o} and \texttt{gpt-4-turbo} are as of 12 July 2024.
    }
    \label{tab:pricing}

\end{table}

\section{Prompts}

Table~\ref{tab:prompts} demonstrates the empirical prompts we use in our experiments as well as in the system demonstration. We use these prompts for \texttt{gpt-4o} and \texttt{gpt-4-turbo} AD translators.

\begin{table}[!htb]
    \centering
    \resizebox{\columnwidth}{!}{
\begin{tabular}{p{\columnwidth}}
\toprule
\textbf{text-only }  \\
\midrule
Translate the following audio description from \{source\_language\} to \{target\_language\}. Respond with the translation only. This is the audio description to translate: \newline  \{audio\_description\}  \\
\midrule
\textbf{text + frames} \\
\midrule
Translate the following audio description for the frames of this video from \{source\_language\} to \{target\_language\}. Respond with the translation only. If the audio description does not match the image, please ignore the image. Respond with a translation only. This is the audio description to translate: \newline \{audio\_description\} \\
\bottomrule
\end{tabular} }
    \caption{Prompts used for translation with \texttt{gpt-4o} and \texttt{gpt-4-turbo}. The placeholders \{source\_language\} and \{target\_language\} denote the respective Swiss languages, while \{audio\_description\} refers to the AD script to be translated. Prompts used for \textbf{text + frames} target both text + 4 frames and text + $n$ frames configurations. The instruction to ignore irrelevant images addresses potential noise from linear sampling.}
    \label{tab:prompts}
\end{table}

\section{Examples}
\label{section:examples}

\begin{figure}[!htb]
\centering
\begin{subfigure}[b]{\columnwidth}

\begin{minipage}{0.5\columnwidth}
        \includegraphics[width=\linewidth]{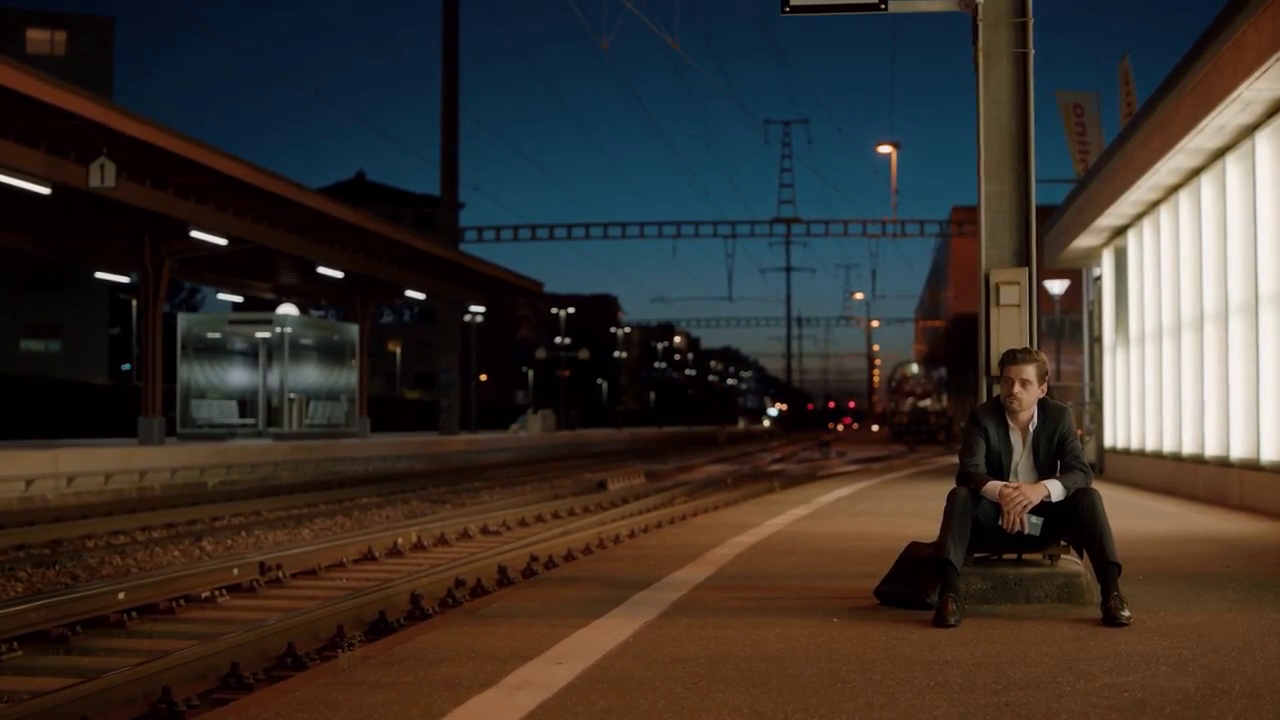}
    \end{minipage}%
    \begin{minipage}{0.5\columnwidth}
        \includegraphics[width=\linewidth]{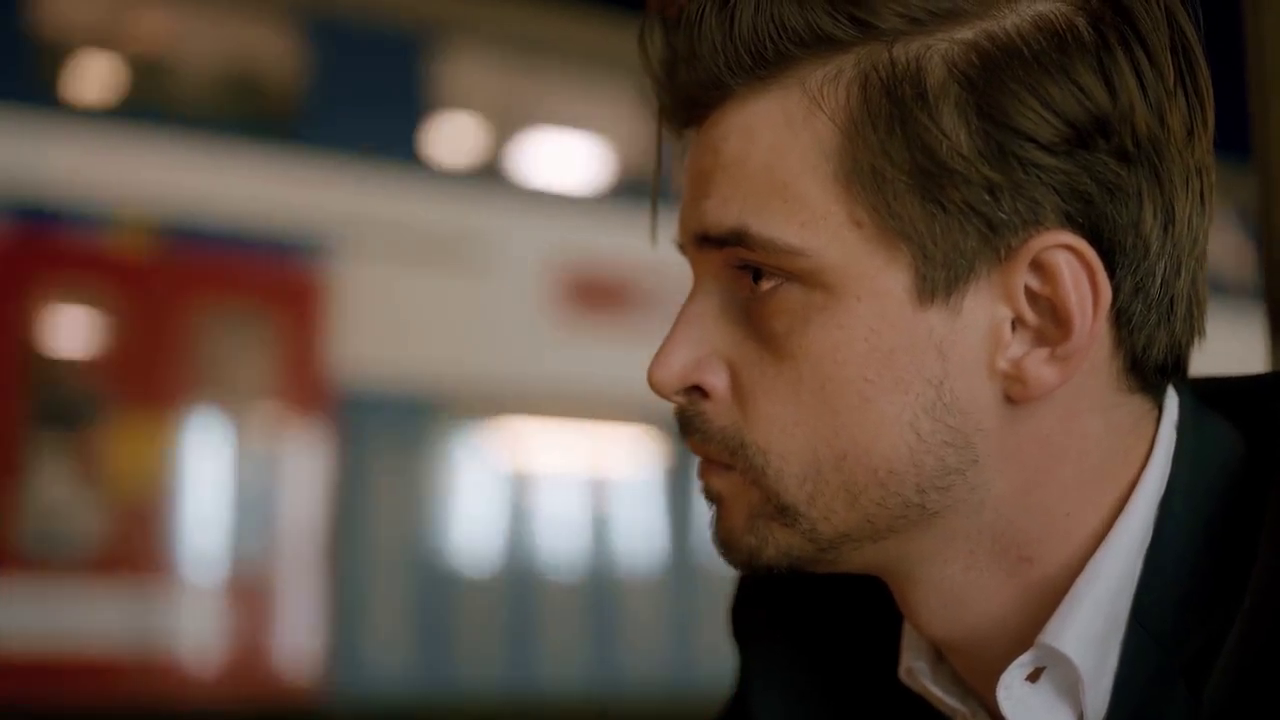}
    \end{minipage}

\caption{Visual context for the AD: \textit{\textbf{Volta} la testa verso un treno che avanza sui binari}. (EN: \textit{\textbf{He} turns \textbf{his} head towards a train moving on the tracks.}) }

\label{fig:examples-train}

\end{subfigure}
\begin{subfigure}[b]{\columnwidth}
\begin{minipage}{0.5\columnwidth}
        \includegraphics[width=\linewidth]{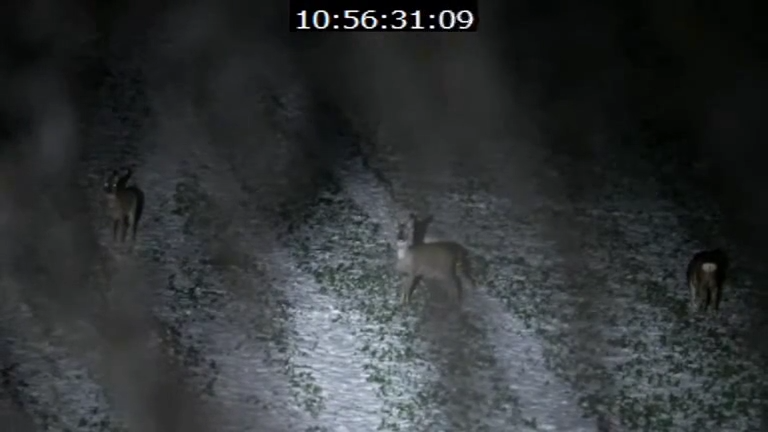}
    \end{minipage}%
    \begin{minipage}{0.5\columnwidth}
        \includegraphics[width=\linewidth]{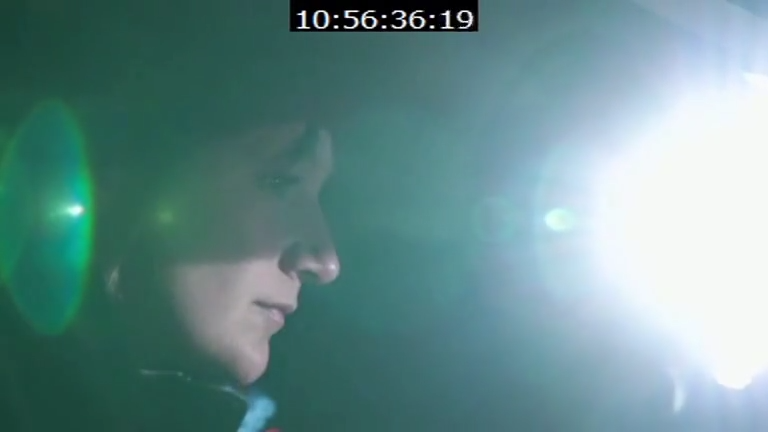}
    \end{minipage}

\caption{Visual context for the AD: \textit{Le \textbf{phare} \'{e}claire deux chevreuils}. (EN: \textit{The \textbf{spotlight} illuminates two deer.}) }

\label{fig:examples-deer}

\end{subfigure}
\caption{Two examples of ambiguity that require additional context for resolution. The words that are correctly disambiguated by the visual input are highlighted in bold.}
\label{fig:examples}
\end{figure}

The following examples demonstrate how multimodal input enhances translation quality by offering extra context. The relevant frames are shown in Figure \ref{fig:examples}.

\paragraph{Grammatical Ambiguity}

The Italian audio description \textit{Volta la testa verso un treno che avanza sui binari} presents multiple translation possibilities. The verb \textit{volta} can be interpreted in two ways:
\begin{itemize}
    \item \textbf{3rd person singular indicative}: \textit{He/she turns his/her head towards a train moving on the tracks.}
    \item \textbf{2nd person singular imperative}: \textit{Turn your head!}
\end{itemize}

This ambiguity is resolved through the visual context of a man sitting on a train platform, as shown in Figure \ref{fig:examples-train}.

\paragraph{Lexical Ambiguity}

The French audio description \textit{Le phare \'{e}claire deux chevreuils} presents two possible translations:
\begin{itemize}
    \item \textit{The lighthouse illuminates two deer.}
    \item \textit{The spotlight illuminates two deer.}
\end{itemize}

The second frame in Figure \ref{fig:examples-deer} clearly shows that, in this context, \textit{phare} should be translated as \textit{spotlight}.

\section{System Demonstration}

\begin{figure}[!htb]
\centering
\begin{subfigure}[b]{\columnwidth}
\includegraphics[width=\textwidth]{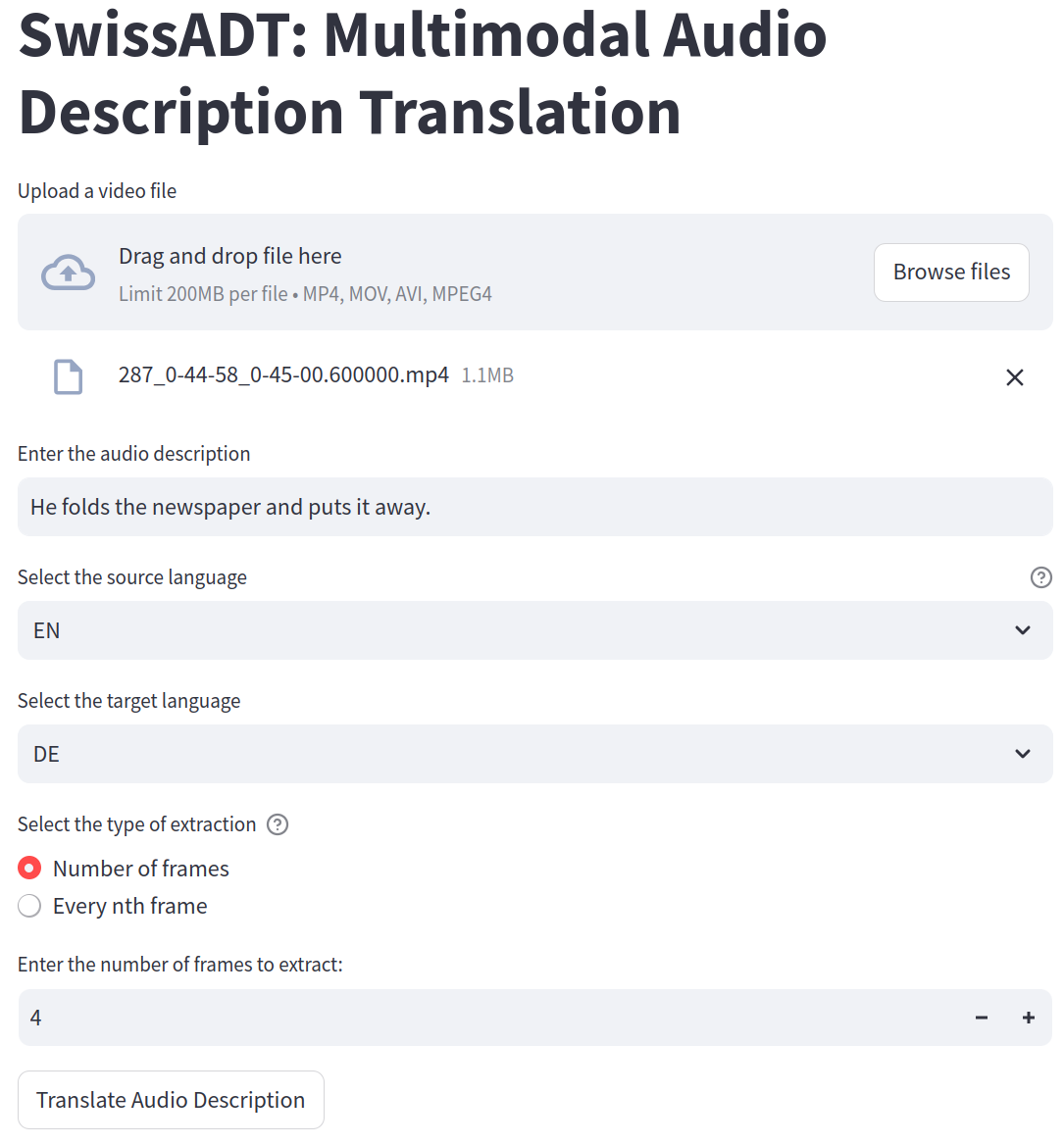}
\caption{\textbf{Demonstration of SwissADT}. To generate the translated AD script from English to German, the user would upload the video clip and provide the AD script in the source language. Additionally, the user would input the number of frames to be sampled from the retrieved moment.}
\label{fig:demo-1}
\end{subfigure}
\begin{subfigure}[b]{\columnwidth}
\includegraphics[width=\textwidth]{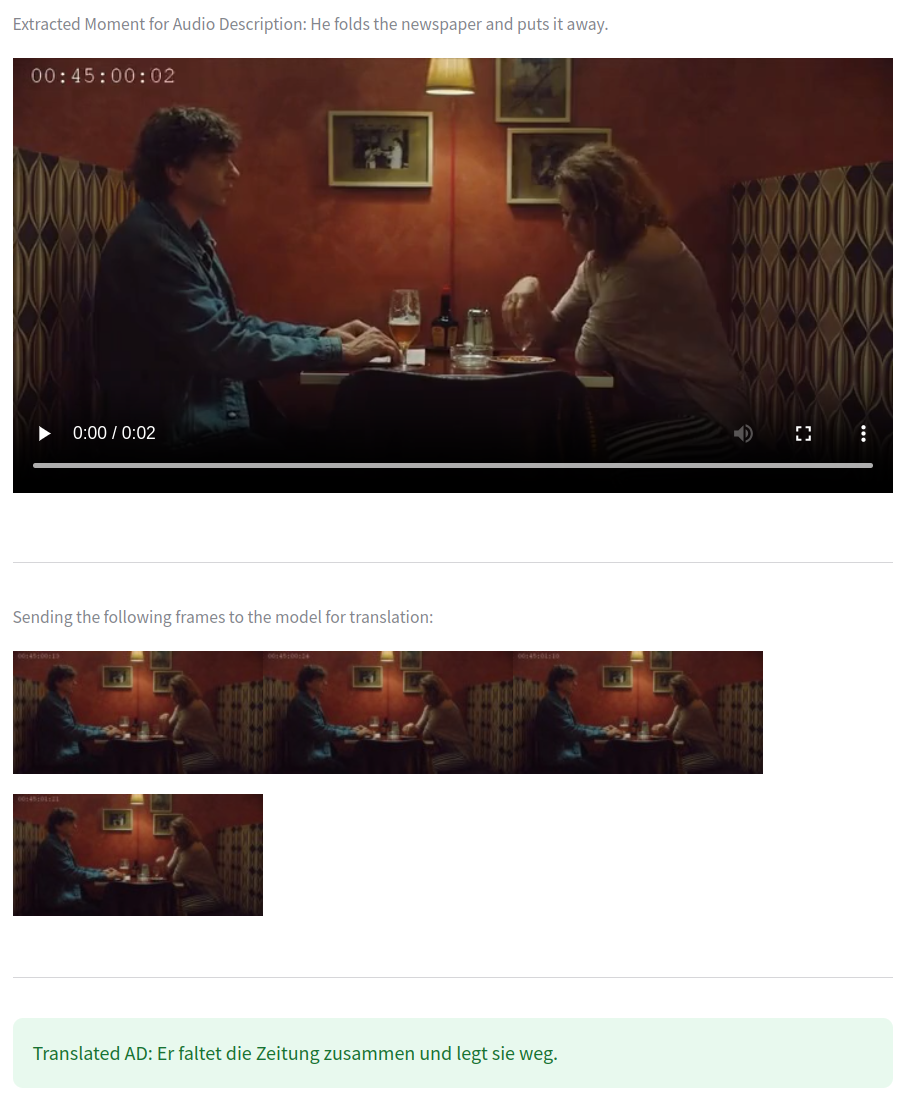}
\caption{\textbf{Generated AD in German}. We display the retrieved moment that best aligned with the source AD script in English, as well as the frames that are linearly sampled from the retrieved moment used by our best AD translator \texttt{gpt-4o}.}
\label{fig:demo-2}
\end{subfigure}
\caption{User interaction interface for SwissADT. We use \texttt{Streamlit} and \texttt{Docker} to implement the user interaction platform.}
\label{fig:demo}
\end{figure}

Our system demonstration for SwissADT (see Figure~\ref{fig:demo} for the system appearance) is hosted at \url{https://github.com/fischerl92/swissADT}. Please follow our detailed instructions on our project page to set up the demo. 

In addition, our demo also runs on our department server at \url{https://pub.cl.uzh.ch/demo/swiss-adt} which can be visited without configurations. We have also recorded a YouTube video explaining how to use the demo, which can be accessed at \url{https://youtu.be/5PQs8DscubU}. 

\end{document}